\documentclass{new_tlp}
\usepackage{amsmath}
\usepackage{graphicx}
\usepackage{multirow}
\usepackage{amsmath}
\newtheorem{examplethm}{Example}
\newtheorem{definition}{Definition}
\newenvironment{example}{\begin{examplethm}\normalfont}{\end{examplethm}}
\usepackage{tikz}
\usetikzlibrary{arrows,positioning,automata,fit}
\usepackage{amsmath}
\usepackage[colorlinks=true]{hyperref} 
\def\st{\medskip\noindent}
\def\no{\; {not} \;}
\newcommand{\rif}{\stackrel{\,\,+}{\leftarrow}}

\newcounter{cm}

\begin{document}

\jdate{March 2003}
\pubyear{2003}
\pagerange{\pageref{firstpage}--\pageref{lastpage}}
\doi{S1471068401001193}

\title[Embracing Background Knowledge in the Analysis of Actual Causality]{Embracing Background Knowledge in the Analysis of Actual Causality:\\An Answer Set Programming Approach}

\author[Michael Gelfond, Jorge Fandinno and Evgenii Balai] {Michael Gelfond\\
Texas Tech University
\and Jorge Fandinno \\
University of Nebraska at Omaha
\and Evgenii Balai \\
Texas Tech University}

\submitted{5 February 2003}
\revised{6 April 2023}
\accepted{11 May 2023}

\maketitle

\begin{abstract}
   This paper presents a rich knowledge representation language aimed at formalizing causal knowledge.
    This language is used for accurately and directly formalizing common benchmark examples from the literature of actual causality.
    A definition of cause is presented and used to analyze the actual causes of changes with respect to sequences of actions representing those  examples.
\end{abstract}


\section{Introduction}
This work is a part of larger research program, originated by John McCarthy and
others in the late fifties.
The program is aimed at the development of Knowledge Representation~(KR) languages capable of clear and succinct formalization of commonsense knowledge.
In this paper we concentrate on a long standing problem of giving a formal account of the notion of actual causality.
Despite significant amount of work in this area the problem remains unsolved.
We believe that the difficulty is related to insufficient attentions paid to relevant commonsense background knowledge.
To analyze causal relations involved in a sequence of events happening even in comparatively simple domains, we need to be able to represent
sophisticated causal laws, time, defaults and their exceptions, recursive definitions, and other non-trivial phenomena of natural language.
To the best of our knowledge none of the KR-languages used in previous works are capable of representing all of these phenomena.
We propose to remedy this problem by analyzing causality in the context of a new rich KR\nobreakdash-language~$\mathcal{W}$ based on the ideas from
Answer Set Prolog~(ASP), Theories of Action and Change~(TAC), and Pearl's  do-operator~\mbox{\cite{pearl2009causality}}.
The language is used to define several causal relations capable of accurate analysis of a number of examples which could not have been properly
analyzed by the previous approaches.
Special emphasis in our approach is given to accuracy and \emph{elaboration tolerance}~\cite{mccarthy98a} of translations of English texts into theories of~$\mathcal{W}$.
This is facilitated by the well developed methodology of such translations in ASP and TAC.
These issues were not typically addressed in work on causality, but they are essential from the standpoint of KR.
We focus on the suitability of~$\mathcal{W}$ for causal analysis, illustrated by its application to well\nobreakdash-known benchmarks from the literature.
The paper is organized as follows.
In the next section we motivate the need for a richer KR\nobreakdash-language by
analyzing such benchmarks.
%
After that, we introduce \emph{causal theories} of~$\mathcal{W}$ and a methodology for formalizing natural language stories.
This is illustrated on these benchmarks. 
Special care is taken of obtaining accurate and direct translation from natural language sentences, and the elaboration tolerance of the representation.
The later is obtained by a clear separation between background commonsense
knowledge (formalized in a \emph{background theory}) and the particular story (formalized as a sequence of events that we call~\emph{scenario}).
Finally, we introduce our definition of cause and discuss several variations of the benchmark examples.
This definition provides answers that match our intuition.
Note that, since $\mathcal{W}$ is a powerful action language
  based on ASP it can also be used
  for reasoning about temporal prediction, planning, etc.
Due to space limitation the paper does not demonstrate the full power of~$\mathcal{W}$ and the full variety of its causal relations.
This will be done in a longer version of the paper.

\section{Motivating Examples}
In this section, we discuss two problematic benchmarks from
the literature and provide their causal description based on KR perspective.
%
We start by considering the \emph{Suzy First} example introduced by~cite{hall04a} and extensively discussed in the literature.
The following reading is by~cite{halpea01a}.

\begin{example}\label{ex:suzy.first}
Suzy and Billy both pick up rocks and throw them at a bottle.
Suzy's rock gets there first, shattering the bottle.
\emph{Since both throws are perfectly accurate, Billy's would have shattered the bottle had it not been preempted by Suzy's throw.
Common sense suggests that Suzy's throw is the cause of the shattering, but Billy's is~not.}
\end{example}

Time and actions are essential features of this example.
The reasoning leading to Suzy's throw being regarded as the cause of the bottle directly points to the sentence ``Suzy's rock gets there first, shattering the bottle.''
Had Billy's throw got there first, we would have concluded that Billy's throw was the cause.
%
%
Despite the importance of time in this example, most approaches do not explicitly represent time.
As a result, the fact that ``Suzy's rock gets there first,'' which naturally is part of the particular scenario, is represented as part of background knowledge~\cite{halpea01a,hoppea03a,chohal04a,hall04a,hall07a,vennekens11a,halpern14a,halpern15a,halpern11a,becven16a,becven18a,bochman18a,denecker2019explaining,beckers2021counterfactual}.
This means that a small change in the scenario such as replacing ``Suzy's rock gets there first, shattering the bottle'' by ``Billy's rock gets there first, shattering the bottle'' or ``Suzy's rock gets there first, but her throw was too weak to shatter the bottle'' requires a complete change of the formal model of the domain instead of a small change to the scenario.
This is a problem of \emph{elaboration tolerance}.
Several approaches addressed the lack of representation of time by introducing features from the area of~\emph{Reasoning about Actions and Change}.
Approaches in the context of the~\emph{Situation Calculus}~\cite{hoppea07a,batsou18a} and~\emph{Logic Programming}~\cite{cabfan14a,cabfan16a,fandinno16a,lebajo19a} allow us to reason about actual causes with respect to different sequences of actions, where the order of these actions matter.
For instance, cite{cabfan16a} explicitly represent a
variation of this example where ``Suzy's rock gets there first'' is replaced by ``Suzy throws first.''
The model associated with this example is represented by the following rules:
\begin{align}
  \mathit{broken}(I) &\leftarrow \mathit{throw}(A,I-1),\, \mathit{not}\ \mathit{broken}(I-1)
  \label{eq:cf.1}
  \\
  \mathit{broken}(I) &\leftarrow \mathit{broken}(I-1),\, \mathit{not}\ \neg \mathit{broken}(I)
  \label{eq:cf.2}
  \\
  \neg\mathit{broken}(I) &\leftarrow \neg\mathit{broken}(I-1),\, \mathit{not}\  \mathit{broken}(I)
  \label{eq:cf.3}
\end{align}
This can be used together with facts:   $\ \neg\mathit{broken}(0), \
  \mathit{throw}(\mathit{suzy},0),\
  \mathit{throw}(\mathit{billy},1)$\\
representing the particular scenario.
%
We can represent an alternative story where ``Billy throws first'' by replacing the last two facts~by
$\mathit{throw}(\mathit{billy},0)$
and
$\mathit{throw}(\mathit{suzy},1)$.
Clearly, this constitutes a separation between model and scenario because
we do not need to modify the rules that represent the model of the domain to accommodate the new scenario.
%
We go a step further and show how to represent the fact that ``Suzy's rock gets there first'' independently of who throws first.
The rock may get there first because Suzy throws first, because she was closer,
etc.  
The reason why her rock gets there first is not stated in the example and it is unnecessary to determine the cause of the shattering.
We are able to do that thanks to the introduction of \emph{abstract time\nobreakdash-steps} in our language, a feature missing in all previously discussed approaches.
%
%
As a second example, consider the \emph{Engineer} scenario introduced by~cite{hall00a}.

\begin{example}\label{engineer}
An engineer is standing by a switch in the railroad tracks. A train approaches in the distance. She flips the switch, so that the train travels down the right\nobreakdash-hand track, instead of the left.
Since the tracks reconverge up ahead, the train arrives at its destination all
the same; let us further suppose that the time and manner of its arrival are exactly as they would have been, had she not flipped the switch.
\end{example}

It is commonly discussed whether flipping the switch should be (part of) the cause of the train arriving at its destination~\cite{halpea01a,hall07a,halpern15a,cabfan16a,becvenn17a,batsou18a}.
%
%
Normally these solutions are  not elaboration tolerant.
For instance, adding a neutral switch position or a third route that does not reconverge, requires a different model or leads to completely different answers.
%


\section{Causal Theory}
This section introduces a simplified version of knowledge representation language $\mathcal{W}$, which is used for the analysis of basic causal
relations. Formally, $\mathcal{W}$ is a \emph{subset} of P-log
  \cite{BaralGR09,BalaiGZ19} expanded by a simple form of constraints with
  signature tailored toward reasoning about change.
Theories of $\mathcal{W}$ are called \emph{causal}.
A causal theory 
consists of a \emph{background theory}~$\mathcal{T}$ representing \emph{general
  knowledge} about the agent's domain and \emph{domain scenario}
$\mathcal{S}$ containing the record of deliberate actions performed by the agents.
A sorted signature $\Sigma$ of $\mathcal{T}$, referred to as
\emph{causal}, consists of 
sorts, object constants, and
function symbols.
Each object constant comes together with its sort;
each function symbol -- with sorts of its parameters and values. In addition to \emph{domain specific sorts} and \emph{predefined sorts} such as
\emph{Boolean}, \emph{integer}, etc., a causal signature
includes sorts for \emph{time\nobreakdash-steps}, \emph{fluents}, \emph{actions}, and
\emph{statics}. Fluents are divided into \emph{inertial}, \emph{transient} and \emph{time\nobreakdash-independent}. %
 An inertial fluent can only change its
value as a result of an action. Otherwise the value remains unchanged. The default value of a transient fluent is \emph{undefined}.
A time\nobreakdash-independent fluent does not depend on time. But, different from  a static, it may change its value after the scenario is expanded by new information.
The value sort of actions is Boolean. 
Terms of~$\Sigma$ are defined as usual.
Let $e$ be a function symbol, $\bar{t}$ be a sequence of ground terms not
containing time-steps, $i$ be a time-step, and $y \in \mathit{range}(e)$.
A ground atom of $\Sigma$ is an expression of one of the forms
\small
$$e(\bar{t},i)=y\  \qquad e(\bar{t},i)\not=y\footnote{As in 
  cite{balduccini2012answer} and cite{BalaiGZ19}, 
  $f(\bar{x}) \not= y$ holds if $f(\bar{x})=z$ such that $z \neq y$.
}
$$
\normalsize
where $e$ is an action, a fluent or a static.
If $e$ does not depend on time, $i$ will be omitted.
%
If $e$ is a Boolean fluent then
$e(\bar{t}) = \top$ (resp. $e(\bar{t}) = \bot$) is sometimes written as $e(i)$
(resp. $\neg e(i)$).
%
%
%
%
%
Atoms formed by actions are called \emph{action atoms}.
Similarly for statics, fluents, etc. Action atom $a(i)$ may be written as $occurs(a,i)$.
%
The main construct used to form background theories of~$\mathcal{W}$ is
 \emph{causal mechanism} (or \emph{causal law}) -- a rule of the form:
\small
\begin{align}\label{eq1}
m :\ e(\bar{t},I)=y  \leftarrow  body, \neg \mathit{ab}(m,I)
\end{align}
\normalsize
where
$e$ is a non\nobreakdash-static, $I$ ranges over time\nobreakdash-steps, $m$ is the unique name of this causal
mechanism, $body$ is a set of atoms of~$\Sigma$ and arithmetic atoms of the
form~$N \prec AE$ where $N$ is a variable or a natural number and
$AE$ is an arithmetic function built from $+$, $-$, $\times$, etc., and~$\prec$
is $=$, $>$, or~$\geq$.
Special Boolean function~$\mathit{ab}(m,I)$ is used to capture exceptions to
application of causal mechanism~$m$ at step $I$.
As usual in logic programming we view causal mechanisms with variables as sets
of their ground instances obtained by replacing variables by their possible
values and evaluating the remaining arithmetic terms.
%
%
%
If $e$ in rule~\eqref{eq1} is an action we refer to $m$ as a \emph{trigger}.
%
A causal mechanism of the form \eqref{eq1} says that ``at time-step $I$,  \emph{body} activates causal mechanism $m$ which, unless otherwise specified, sets the value of~$e$ to~$y$''.
%
%
%
\addtocounter{cm}{1} 
\begin{figure*}
\small
\centering
$$
\begin{array}{lclll}
m_1 &: &
\mathit{arrived}(\mathit{fork}) & \leftarrow &
      \mathit{approach}(I),\
      \neg \mathit{ab}(m_1,I)
\\[2pt]
m_2 &: &
\mathit{arrivTime}(\mathit{fork}) = I  & \leftarrow &
      \mathit{arrived}(\mathit{fork}),\
      \mathit{approach}(I-\mathit{time2fork}),\
      \neg \mathit{ab}(m_2,I)
\\[2pt]
m_3(P) &: & \mathit{switch}(I) = P & \leftarrow & flipTo(P,I-1),\ \mathit{switch}(I-1) \neq P,\, \neg \mathit{ab}(m_3(P),I)
\\[2pt]
m_4 &: &
    \mathit{arrived}(\mathit{dest}) & \leftarrow &
    \mathit{arrivTime}(\mathit{fork}) = I,\
    \mathit{switch}(I) \neq \mathit{neutral},\
    \neg \mathit{ab}(m_4,I)
\\[2pt]
m_5 &: &
    \mathit{arrivTime}(\mathit{dest}) = I & \leftarrow &
    \mathit{arrived}(\mathit{dest}),\
    \mathit{arrivTime}(\mathit{fork}) = I',\
    \mathit{switch}(I') = P,\\
    &&&&I = I' + \mathit{time2dest}(P),\
    \neg \mathit{ab}(m_5,I)
\end{array}
$$
\normalsize
\caption{Causal mechanisms in the background theory representing the Engineer story}
\label{fig:eng}
\end{figure*}%
To conform to this reading  we need to enforce a broadly shared \emph{principle of
causality: ``the cause must precede its effect''}. Our version of this
principle is given by the
following requirement: For every ground instance of causal mechanism such that~$i$ is a time\nobreakdash-step occurring in its head and~$j$ is a  time\nobreakdash-step occurring in its body, the following two conditions are satisfied:
\begin{itemize}
\item $j < i$ if $j$ occurs within an action atom; and
\item $j \leq i$, otherwise.
\end{itemize}
A \emph{scenario} of background theory $T$ with signature $\Sigma$ is a
collection of static and arithmetic atoms together with \emph{expressions} of the form:
\begin{itemize}
\item ${\bf init}(f=y)$ -- the initial value of
  inertial fluent $f$ 
    is $y$;
\item ${\bf do}(a,i)$ -- an agent deliberately executes action $a$ at
time\nobreakdash-step $i$;
\item ${\bf do}(\neg a,i)$ -- an agent deliberately refrains from
  executing action $a$ at $i$;
  \item
  ${\bf obs}(f,y,i)$ -- the value of $f$ at time-step $i$
  is observed to be $y$;

\end{itemize}
We refer to these expressions as \emph{extended atoms of~$\Sigma$};
a set of extended atoms of the form~${{\bf init}(f=y)}$, ${{\bf
    init}(g=z)}\dots$ will be written as~${{\bf init}(f=y,\, g=z, \dots)}$.
%
We assume that the sort for time\nobreakdash-steps consists of all natural
numbers and symbolic constants we refer to as \emph{abstract time-steps}.
Atoms, extended atoms and scenarios where all object constants of the sort for
time\nobreakdash-steps are natural numbers are called \emph{concrete}; those
that contain abstract time-steps are called~\emph{abstract}.

The story of~\emph{Suzy First} (Example~\ref{ex:suzy.first}) can be represented
in $\mathcal{W}$ by a background theory~$\mathcal{T}_{\mathit{fst}}$ which contains a sub-sort $\mathit{throw}$ of actions,
inertial fluent $\mathit{broken}$, statics $\mathit{member}$, $\mathit{agent}$, and
$\mathit{duration}$ and causal mechanism
\small
\begin{gather*}
\begin{array}{rlll}
m_{0}(A) : & \mathit{broken}(I) &\leftarrow &
         \mathit{occurs}(A,I-D),\mathit{member}(A,throw),\\
     &&&\mathit{agent}(A) = Ag,\mathit{duration}(A)=D,\\
     &&& \neg \mathit{broken}(I-1),\neg \mathit{ab}(m_0(A),I)
\end{array}
\end{gather*}%
\normalsize
The theory will be used together with an abstract scenario
$\mathcal{S}_{\mathit{suzy}}$ which includes
actions $a_1$ and $a_2$ of the sort $\mathit{throw}$ and
atoms
\small
\begin{gather*}
{\bf init}(\neg broken),\ {\bf do}(a_1,t_1),\ {\bf
  do}(a_2,t_2),\ t_1 \!+\! \mathit{duration}(a_1) < t_2 \!+\! \mathit{duration}(a_2)
\end{gather*}
\normalsize
where~$t_1$ and~$t_2$ are abstract time\nobreakdash-steps.
The last (arithmetic) atom represents the fact that Suzy's stone arrives first.
Actions of  $\mathcal{S}_{\mathit{suzy}}$ are
described by statics
\small
\begin{flalign}
    \label{eq:suzy.statics}
\quad&\begin{aligned}
  \mathit{agent}(a_1)=suzy \quad \mathit{member}(a_1,throw)\quad
  \mathit{agent}(a_2)=billy \quad \mathit{member}(a_2,throw)
\end{aligned}&
\end{flalign}
\normalsize
and arithmetic atoms $\mathit{duration}(a_1)\geq 1$, $\mathit{duration}(a_2) \geq 1$.
Here, and in other places $f(\bar{t}) \geq y$ is understood as a shorthand
for
$f(\bar{t}) = d$ and $d \geq y$, where $d$ is a fresh abstract constant.
Similarly for $>$, $=$ and $\not=$.
%
To save space, we omit executability conditions for causal mechanisms.
Note that causal mechanism~$m_0(A)$ is a general commonsense law which is not specific to this particular story.
This kind of general commonsense knowledge can be compiled into a background library and retrieved when necessary~\cite{inclezan16a}.
The same applies to all the other causal mechanisms for variations of this example discussed in the paper.
%
%
Note that we explicitly represent the temporal relation among time\nobreakdash-steps and \emph{make no further assumptions} about the causal relation among the rocks.
The definition of cause introduced below is able to conclude that Suzy's rock is the cause of breaking the bottle. 
This is a distinguishing feature of our approach.
%
Representing that Billy's stone arrives first is obtained simply by replacing the corresponding constraint in  $\mathcal{S}_{\mathit{suzy}}$  by ~$t_1 + \mathit{duration}(a_1) > t_2 + \mathit{duration}(a_2)$.

\st
The \emph{Engineer} story (Example~\ref{engineer}) can be represented 
 by a background theory~$\mathcal{T}_{\mathit{eng}}$ containing causal mechanisms in Figure~\ref{fig:eng}.
%
%
The arrival of the train is modeled by a time\nobreakdash-independent
fluent~$\mathit{arrived}(\mathit{point})$. Action
$\mathit{approach}$ of $m_2$ causes the train to arrive at the fork after the amount of
time determined by static~$\mathit{time2fork}$ (note that since $m_1$
can fail to cause $arrived(fork)$, this atom cannot be removed from $m_2$).
The switch 
is controlled by action $\mathit{flipTo}$ which takes one unit of time. This
action can change the switch to any of its three positions: $\mathit{neutral}$,
$\mathit{left}$, and $\mathit{right}$.
Static~$\mathit{time2dest}(\mathit{track})$ determines the time it takes the
train to traverse the distance between the fork and the train's destination
depending on the $\mathit{track}$ taken. When the switch is in the
$\mathit{neutral}$ position, the train does not arrive at its destination.
Inertial fluent~$\mathit{switch}$ represents the position of the switch.
%
%
The times to travel between two points must obey the following constraints included in scenario~$\mathcal{S}_{eng}$:
\small
\begin{gather*}
\mathit{time2fork} \geq 1 \quad
\mathit{time2dest}(\mathit{left}) \geq 1 \quad
\mathit{time2dest}(\mathit{right}) \geq 1
\end{gather*}
\normalsize
The rest of the scenario~$\mathcal{S}_{eng}$ consists of the following atoms
\small
\begin{gather*}
{\bf init}(\mathit{switch}=left)\ \quad
{\bf do}(\mathit{approach},t_3)\ \quad
{\bf do}(\mathit{flipTo(right)},t_4)\quad\\
\mathit{time2dest}(\mathit{left}) = \mathit{time2dest}(\mathit{right})
\end{gather*}
\normalsize
We make no assumptions regarding the order in which actions~$\mathit{approach}$
and~$\mathit{flipTo}$ occur.
We can easily modify the scenario to accommodate a variation of the story where traveling down the right\nobreakdash-hand track is faster than over the left one by replacing the last arithmetic atom by~$\mathit{time2dest}(\mathit{left}) > \mathit{time2dest}(\mathit{right})$.

\begin{definition}[Causal Theory]\label{causalTheory}
A \emph{causal theory} $\mathcal{T}(\mathcal{S})$ is a pair where~$\mathcal{T}$ is a background theory and $\mathcal{S}$ is a scenario.
\end{definition}

\noindent
We identify each causal theory $\mathcal{T}(\mathcal{S})$ with the logic program
that consists of causal mechanisms
without their labels,
all atoms in~$\mathcal{S}$ as facts and the following general axioms:
\small
\begin{align}
    \label{eq:axiom.det1}
\mathit{def}(f(\bar{X})) \leftarrow f(\bar{X}) = Y
\end{align}
\small
\begin{align}
    \label{eq:axiom.det2}
\leftarrow f(\bar{X}) \neq Y, \no \mathit{def}(f(\bar{X}))
\end{align}
\small
\begin{align}
    \label{eq:axiom.neq}
f(\bar{X}) \neq Y \leftarrow f(\bar{X}) = Z,\, Z \neq Y
\end{align}
for every function symbol~$f$,
\normalsize
\small
\begin{align}
\label{eq:ab}
\neg \mathit{ab}(m,I) &\leftarrow\no \mathit{ab}(m,I)
\end{align}
\normalsize
for every causal mechanism~$m$,
\small
\begin{align}
    \label{eq:init}
    f(0)=y \leftarrow {\bf init}(f=y)
\end{align}
\normalsize
\small
\begin{align}
\label{eq:inertia}
f(I)=Y &\leftarrow  f(I - 1)=Y,\no f(I)\neq Y
\\
\label{eq:inertia.neq}
f(I) \neq Y &\leftarrow  f(I - 1) \neq Y,\no f(I) = Y
\end{align}
\normalsize
for every inertial fluent~$f$,
\small
\begin{align}
    \label{eq:action.pos}
    a(I) &\leftarrow {\bf do}(a,I)\ \ \
    & \neg a(I) &\leftarrow {\bf do}(\neg a,I)
    \\
    \label{eq:action.default}
  \neg a(I) &\leftarrow \no a(I)
\\
    \label{eq:action.cr}
a(I) &\rif
    \\
    \label{eq:action.ab}
    \mathit{ab}(m,I) &\leftarrow {\bf do}(a = v,I)
\end{align}
\normalsize
for every action~$a$, Boolean value~$v$ and causal law~$m$ with head~$a(I) = w$ and~$v \neq w$.
 \begin{align}
     \label{eq:axiom.obs}
     \leftarrow obs(f(\bar{X}),Y,I), not~f(\bar{X},I) = Y
 \end{align}
%
Axioms~\eqref{eq:axiom.det1}, \eqref{eq:axiom.det2}, and \eqref{eq:axiom.neq}
reflect the reading of relation $\neq$.
Axiom~\eqref{eq:ab} ensures that causal mechanisms are defeasible.
%
Axiom~\eqref{eq:init} ensures that fluents at the initial situation take the value described in the scenario.
Axioms~(\ref{eq:inertia}-\ref{eq:inertia.neq}) are the \emph{inertia axioms}, stating that inertial
fluents normally keep their values.
%
Axiom~(\ref{eq:action.pos}) ensures that the actions occur as described in the scenario.
Axiom~\eqref{eq:action.default}
states the close world assumption for actions.
Axiom \eqref{eq:action.cr} is a cr-rule~\cite{balduccini2003logic,gelfond2014knowledge} which allows indirect exceptions to
\eqref{eq:action.default}. Intuitively, it says that \emph{$a(I)$ may be true,
  but such a possibility is very rare and, whenever possible, should be ignored.}
Axiom~\eqref{eq:action.ab} ensures that deliberate actions overrule the default behavior of contradicting causal mechanisms (See Example~\ref{ex:suzy.order} below for more details).
Axiom~\eqref{eq:axiom.obs} ensures that observations are satisfied in the model.
%
%
Note, that if $\mathcal{T}(\mathcal{S})$ contains occurrences of abstract
time-steps then its grounding may still have occurrences of arithmetic
operations.
(If $d$ is an abstract time-step then, say, $d+1 > 5$ will be unchanged by
the grounding).
The standard definition of answer set is not applicable in this
case. The following modification will be used to define the meaning of
programs with abstract time-steps.
Let $\gamma$ be a mapping of abstract time-steps into the natural numbers
and $\mathcal{T}(\mathcal{S})$ be a program not containing variables.
By~$\mathcal{T}(\mathcal{S})|_\gamma$ we denote the result of
\begin{itemize}
  \item[(a)] applying
  $\gamma$ to abstract time-steps from~$\mathcal{T}(\mathcal{S})$,
  \item [(b)] replacing
  arithmetic expressions by their values,
  \item [(c)] removing rules containing false
  arithmetic atoms.
\end{itemize}
Condition (c) is needed to avoid violation of principle of causality by useless rules.
By an \emph{answer set} of~$\mathcal{T}(\mathcal{S})$ we mean an answer set of~$\mathcal{T}(\mathcal{S})|_\gamma$ for some $\gamma$.
If~$\mathcal{T}(\mathcal{S})|_\gamma$ is
consistent, i.e., has an answer set then $\gamma$ is called an
\emph{interpretation} of $\mathcal{T}(\mathcal{S})$. $\mathcal{T}(\mathcal{S})$
is called consistent if it has at least one interpretation.
If $\mathcal{T}(\mathcal{S})$ is consistent and for each interpretation
$\gamma$ of $\mathcal{T}(\mathcal{S})$, $\mathcal{T}(\mathcal{S})|_\gamma$ has
exactly one answer set then $\mathcal{T}(\mathcal{S})$ is called
\emph{deterministic}.
\emph{In this paper we limit ourselves to deterministic causal theories}.
%
%
%
%
To illustrate our representation of triggers, parallel actions, and the defeasibility of causal laws, we introduce the following variation of \emph{Suzy First} (Example~\ref{ex:suzy.first}).
\begin{example}\label{ex:suzy.order}
  Suzy and Billy throw rocks by the order of a stronger girl.
Suzy's rock gets there first.
\end{example}
The effects of orders are described by the causal mechanism:
\small
\begin{gather*}
    \label{eq:order}
\begin{array}{rllll}
  m_6(A,T,B) :&  \mathit{occurs}(A,I) & \leftarrow & \mathit{member}(B,\mathit{order}),& \mathit{occurs}(B,T),\\
            &&& \mathit{what}(B) = A, & \mathit{when}(B)=I,\\
            &&& I > T, & \neg \mathit{ab}(m_6(A,T,B),I).
\end{array}
\end{gather*}
\normalsize
%
%
The scenario~$\mathcal{S}_{order}$ is obtained from~$\mathcal{S}_{suzy}$
by adding new actions, $b_1$ and $b_2$, of the sort~$\mathit{order}$  described
 by statics
\small
\begin{flalign}
    \label{eq:orders.statics}
&\begin{array}{l}
\mathit{what}(b_1)=a_1 \ \ \mathit{when}(b_1) = t_1\ \  \mathit{what}(b_2) = a_2\ \
\mathit{when}(b_2) = t_2
\end{array} 
\end{flalign}
\normalsize
and new constraints  $t_1>0,\ t_2>0,$
and replacing its extended atoms by
%
\small
\begin{gather*}
    {\bf init}(\neg broken), \ {\bf do}(b_1,0), \
    {\bf do}(b_2,0).
\end{gather*}
\normalsize
For any interpretation~$\gamma$, in the unique answer set of~$\mathcal{T}(\mathcal{S}_{order})|_\gamma$ atom $\mathit{broken}$ becomes true
at time\nobreakdash-step%
\footnote{
%
$\mathit{duration}(a_1)$ stands for $\mathit{d}$ where $duration(a_1)=d$.
}
~${\gamma(t_1)+\gamma(\mathit{duration}(a_1))}$.
%
For the sake of simplicity, we assume that orders are given at
time\nobreakdash-step~$0$, but in general we would use two abstract time-steps.
%
The example illustrates representation of triggers and parallel actions.
To illustrate defeasibility, let us consider a scenario where both Suzy and Billy refuse to follow the order.
This can be formalized as scenario~$\mathcal{S}_{order2}$ obtained from~$\mathcal{S}_{order}$ by adding the extended atoms~${\bf do}(\neg a_1,t_1)$ and~${\bf do}(\neg a_2,t_2)$.
Due to axioms~\eqref{eq:action.ab}, causal mechanisms~$m_6(a_1,0,b_1)$
and~$m_6(a_2,0,b_2)$ do not fire, and $\mathit{broken}$ never becomes true.
%
%
\section{Cause of Change}
In this section, we describe our notion of cause of change.
We start with scenarios not containing observations.
\begin{definition}
We say that a ground atom~$e(\bar{t},k) = y$ is a \emph{change} in~$\mathcal{T}(\mathcal{S})|_\gamma$ if the unique answer set~$M$
of~$\mathcal{T}(\mathcal{S})|_\gamma$ satisfies~$e(\bar{t},k) = y$ and one of the
following conditions holds:
\begin{itemize}
  \item $e$ is inertial and~$e(\bar{t},k-1)$ is either undefined in~$M$ or $M$
    satisfies~$e(\bar{t},k-1) = z$ with~$z \neq y$;
  \item $e$ is an action or a transient or time\nobreakdash-independent fluent.
\end{itemize}
\end{definition}
The definition of cause of change  relies on the definition of \emph{tight proof} that we introduce next.
By~$[P]_i$, we denote the sequence consisting of the first~$i$ elements of sequence $P$.
%
By~$\mathit{atoms}(P)$ we denote the atoms occurring in~$P$.

\begin{definition}[Proof]\label{proof}
A \emph{proof} of a set~$\mathcal{U}$ of ground atoms in~$\mathcal{T}(\mathcal{S})|_\gamma$ is a sequence $P$ of atoms in the unique answer set~$M$ of $\mathcal{T}(\mathcal{S})|_\gamma$
and rules of the ground logic program associated with~$\mathcal{T}(\mathcal{S})|_\gamma$ satisfying the following conditions:
\begin{itemize}
\item $P$ contains all the atoms in~$\mathcal{U}$.
\item Each element $x_i$ of $P$ is one of the following:
  \begin{itemize}
  \item a rule whose body is satisfied by the set
    $$atoms([P]_i) \cup \{\mbox{not } l : l \not\in M\}, \text{ or}$$
  \item an axiom, i.e., a {\bf do}-atom or a static from $\mathcal{S}|_\gamma$, or
    \item the head of some rule from $[P]_i$.
\end{itemize}
  \item No proper subsequence%
  \footnote{A sequence obtained from $P$ by removing some of its elements.}
  of $P$ satisfies the above conditions.
\end{itemize}
\end{definition}

Let us consider the \emph{Engineer} story (Example~\ref{engineer}) and an interpretation~$\gamma$ of the abstract theory~$\mathcal{T}_{eng}(\mathcal{S}_{eng})$, that is, a function mapping~$\mathit{time2fork}$, $\mathit{time2dest}(\mathit{left})$ and~$\mathit{time2dest}(\mathit{right})$ to natural numbers such that
\small
$$\gamma(\mathit{time2dest}(\mathit{left})) = \gamma(\mathit{time2dest}(\mathit{right})).$$
\normalsize
For instance, an interpretation~$\gamma$ satisfying~$\gamma(t_3) = 0$,
${\gamma(t_4) = 1}$,
$\gamma(\mathit{time2fork}) = 3$ and
\small
$$\gamma(\mathit{time2dest}(\mathit{left})) =
\gamma(\mathit{time2dest}(\mathit{right})) = 5. $$
\normalsize
The unique answer set of~$\mathcal{T}_{eng}(\mathcal{S}_{eng})|_\gamma$ contains, among others, atoms
\small
\begin{align*}
    \mathit{switch}(0) \neq \mathit{neutral},
    \dots,
                            \mathit{switch}(3) \neq \mathit{neutral}
\end{align*}
    \begin{align*}
    \mathit{arrivTime}(\mathit{fork}) = 3,
    \phantom{\neg} \mathit{arrived}(\mathit{dest})
\end{align*}
\normalsize
Since~$\mathit{arrived}$ is a time\nobreakdash-independent fluent, we can
conclude that~$\mathit{arrived}(\mathit{dest})$ is a change in this concrete
causal theory.
In general, we can check that, for any interpretation~$\gamma$ of the abstract theory~$\mathcal{T}_{eng}(\mathcal{S}_{eng})$ where the switch is flipped before the train arrives to the fork, i.e. satisfying~${\gamma(t_4) < \gamma(t_3) + \gamma(\mathit{time2fork})}$,
the unique answer set of~$\mathcal{T}_{eng}(\mathcal{S}_{eng})|_\gamma$ contains atoms
\small
\begin{align*}
    \mathit{switch}(0) \neq \mathit{neutral},
    \dots,
  \mathit{switch}(n_1)\neq \mathit{neutral}
  \end{align*}
 \begin{align*}
    \mathit{arrivTime}(\mathit{fork}) = n_1,
    \phantom{\neg} \mathit{arrived}(\mathit{dest})
                              \end{align*}
\normalsize
where~${n_1 = \gamma(t_3) + \gamma(\mathit{time2fork})}$
is a natural number corresponding to the arrival time of the train to the switch.
%
We can then conclude that~$\mathit{arrived}(\mathit{dest})$ is a change in this causal theory for any such interpretation~$\gamma$.
\begin{figure}[t]
\centering
\scriptsize
\begin{gather*}
\begin{aligned}[t]
&{\bf do}(\mathit{approach},\gamma(t_3))\\
&m_1 \text{ at $I =$ } \gamma(t_3)\\
&\mathit{arrived}(\mathit{fork})\\
&m_2 \text{ at $I =$ }n_1\\
&\mathit{arrivTime}(\mathit{fork}) = n_1\\
&\mathit{switch}(0) \neq \mathit{neutral}\\
&\dots \quad (\textit{inertia rules})\\
&\mathit{switch}(n_1) \neq \mathit{neutral}\\
&m_4 \text{ at $I =$ }n_1\\
&\mathit{arrived}(\mathit{dest})\\
\end{aligned}
\hspace{1cm}
\begin{aligned}[t]
&{\bf do}(\mathit{approach},\gamma(t_3))\\
&m_1 \text{ at $I =$ }\gamma(t_3)\\
&\mathit{arrived}(\mathit{fork})\\
&m_2 \text{ at $I =$ }n_1\\
&\mathit{arrivTime}(\mathit{fork}) = n_1\\
&{\bf do}(\mathit{flipTo(right)},\gamma(t_4))\\
&m_3(right) \text{ at $I =$ }\gamma(t_4) + 1\\
&\mathit{switch}(\gamma(t_4)+1) = right\\
&axiom \ \eqref{eq:axiom.neq}\\ 
&\mathit{switch}(\gamma(t_4)+1) \neq \mathit{neutral}\\
&\dots \quad (\textit{inertia rules})\\
&\mathit{switch}(n_1) \neq \mathit{neutral}\\
&m_4 \text{ at $I =$ }n_1\\
&\mathit{\mathit{arrived}(\mathit{dest})}\\
\end{aligned}
\end{gather*}
\normalsize
\caption{Proofs~$P_1$ and~$P_2$ of~$\mathit{arrived}(\mathit{dest})$ in scenario~$\mathcal{S}_{eng}$ with any interpretation~$\gamma$ satisfying condition~${\gamma(t_3)+\gamma(\mathit{time2fork}) > \gamma(t_4)}$.
We use~$n_1$ to denote the positive natural number~$\gamma(t_3)+\gamma(\mathit{time2fork})$.
Only~$P_1$ is a tight proof.
}
\label{fig:train.proofs.s1}
\end{figure}
Figure~\ref{fig:train.proofs.s1} depicts (condensed versions) of the two proofs in this scenario for any such interpretation~$\gamma$.
In~$P_1$, we reach the conclusion that the switch is not in the neutral position by inertia.
In~$P_2$, the same conclusion is the result of the flipping the switch to the $\mathit{right}$.
Both are valid  derivations of the change~${\mathit{arrived}(\mathit{dest})}$.
However, to infer the causes of an event we give preference to proofs using inertia over those using extra causal mechanisms.
This idea is formalized in the following notion of~\emph{tight proof}.
\begin{definition}[Tight proof]
  Let~$P_1$ and~$P_2$ be proofs of change~$e(\bar{t},k)=y$ in
  $\mathcal{T}(\mathcal{S})|_\gamma$. $P_1$ is \emph{(causally) tighter}
  than~$P_2$ if every causal mechanism of~$P_1$ belongs to~$P_2$ but not
  vice\nobreakdash-versa.
Proof~$P$ of $e(\bar{t},k)=y$ in~$\mathcal{T}(\mathcal{S})|_\gamma$ is \emph{tight} if there is no proof of $e(\bar{t},k)=y$ in~$\mathcal{T}(\mathcal{S})|_\gamma$ that is tighter than~$P$.%
\end{definition}
Clearly, proof~$P_1$ from our running example is tighter
than~$P_2$; causal mechanisms of~$P_1$ are $m_1$, $m_2$ and~$m_4$, while~$P_2$
contains the additional causal mechanism $m_3(right)$.
\begin{definition}[Causal chain]\label{causalChain}
Given a numeric time\nobreakdash-step~$i$ and an atom~${e(\bar{t},k)=y}$ in $\mathcal{T}(\mathcal{S})|_\gamma$,
a \emph{causal chain}~$Ch(i)$ from~$i$ to~$e(\bar{t},k)=y$ is a sequence
$$a_1,\ldots,a_n, C_1,\ldots,C_m, e(\bar{t},k) = y$$
of atoms and ground causal mechanisms of~$\mathcal{T}(\mathcal{S})|_\gamma$ with~${n \geq 1}$ and~$m \geq 0$ such that there is a tight proof~$P$ of $e(\bar{t},k)=y$ in~$\mathcal{T}(\mathcal{S})|_\gamma$ satisfying the following conditions:
\begin{itemize}
\item $a_1$ is a do-atom from $P$ with time step $i$,
\item $a_2,\ldots a_n$ are all other do-atoms from $P$ with time-steps greater than or equal to $i$, and
\item $C_1,\ldots,C_m$ are all causal mechanisms of $P$ with time-steps greater than $i$.
\end{itemize}
\end{definition}
Let us introduce some terminology.
We say that $Ch(i)$ is \emph{generated} from the proof $P$ above.
If~$e(\bar{t},k)=y$ is a change, we say that causal chain from~$i$ to ${e(\bar{t},k)=y}$ in $\mathcal{T}(\mathcal{S})|_\gamma$ \emph{leads to change~$e(\bar{t},k)=y$}.
A causal chain is \emph{initiated} by the set of all its do\nobreakdash-atoms.
Two proofs of a set of ground atoms~$\mathcal{U}$ are \emph{equivalent} if they differ only by the
order of their elements.
Two chains are \emph{equivalent} if they are generated from equivalent
proofs.

Continuing with our running example, sequence
\small
\begin{gather}
    \label{eq:arrived.causal.chain}
    {\bf do}(\mathit{approach},\gamma(t_3)),\ m_1, \ m_2,\ \ m_4, \ \mathit{arrived}(\mathit{dest})
\end{gather}
\normalsize
is a causal chain in this scenario that leads to change ${\mathit{\mathit{arrived}(\mathit{dest}})}$, and it is generated by proof~$P_1$ in Figure~\ref{fig:train.proofs.s1}.
However, sequence
\small
\begin{gather*}
   {\bf do}(\mathit{approach},\gamma(t_3)),\ {\bf do}(\mathit{flipTo(right)},\gamma(t_4)),\ m_1, \ m_2,\\ m_3(right), \ m_4, \ \mathit{arrived}(\mathit{dest}  )
\end{gather*}
\normalsize
corresponding to proof~$P_2$ is not causal chain because~$P_2$ is not a tight proof.

\begin{definition}[More informative causal chain]\label{InformativeCausalChain}
  Given causal chains $Ch(i)$ and $Ch(j)$ 
  to~$e(\bar{t},k)=y$,
we say that $Ch(i)$ \textit{is more informative than} $Ch(j)$  if $i < j$ and
$Ch(i)$ contains all elements of~$Ch(j)$.
\end{definition}
%
\begin{definition}[Candidate inflection point]\label{candidateInflectionPoint}
A time-step $i$ is called a \emph{candidate inflection point of change~$e(\bar{t},k) = y$ in $\mathcal{T}(\mathcal{S})|_\gamma$} if
it satisfies the following conditions:
\begin{itemize}
\item [(a)] There is a causal chain from $i$ to~$e(\bar{t},k)=y$ in
 $\mathcal{T}(\mathcal{S}[i])|_\gamma$, and
 \item [(b)] There is a causal chain from $i$ to $e(\bar{t},k)=y$ in
   $\mathcal{T}(\mathcal{S})|_\gamma$
\end{itemize}
where~$\mathcal{S}[i]$ is the scenario obtained from~$\mathcal{S}$ by removing all {\bf do}-atoms after~$i$.
\end{definition}
\begin{definition}[Inflection point]\label{inflectionPoint}
A candidate inflection point $i$ is called an \emph{inflection point of~$e(\bar{t},k) = y$ in $\mathcal{T}(\mathcal{S})|_\gamma$} if there is a causal chain $Ch(i)$ from $i$ to~$e(\bar{t},k) = y$
in $\mathcal{T}(\mathcal{S})|_\gamma$ such that there is no candidate inflection point $j$ and causal
chain $Ch(j)$ from $j$ to $e(\bar{t},k)=y$ in $\mathcal{T}(\mathcal{S})|_\gamma$ which
is more informative than $Ch(i)$.
\end{definition}

Note that a scenario can have more than one inflection point
(see Example~\ref{ex:suzy.same}).
\begin{definition}[Deliberate cause of change]\label{causeOfChange}
A non-empty set $\alpha$ of do\nobreakdash-atoms is called a \emph{(deliberate) cause of change}~${e(\bar{t},k)=y}$ in~$\mathcal{T}(\mathcal{S})|_\gamma$ if
there is an inflection point~$i$ of~${e(\bar{t},k)=y}$ in~$\mathcal{T}(\mathcal{S})|_\gamma$ such that~$\alpha$ initiates a causal chain in~$\mathcal{T}(\mathcal{S})|_\gamma$ from $i$ to ${e(\bar{t},k)=y}$.

It is said to be \emph{(deliberate) cause of change}~${e(\bar{t},k)=y}$ in~$\mathcal{T}(\mathcal{S})$ if it is a cause of change~${e(\gamma(k))=y}$ in~$\mathcal{T}(\mathcal{S})|_\gamma$ for every interpretation~$\gamma$ of~$\mathcal{T}(\mathcal{S})$.
\end{definition}

Following with the Engineer example (Example~\ref{engineer}),
let us consider scenario~$\mathcal{S}_{eng}$.
Since this is an abstract scenario, to answer questions about the cause of change
we have to consider all interpretations of this scenario.
We proceed by cases.
Let us first consider an interpretation~$\gamma$
satisfying condition~${\gamma(t_4) < \gamma(t_3) +
\gamma(\mathit{time2fork})}$.
As we discussed above, \eqref{eq:arrived.causal.chain} is the only causal chain leading to change~${\mathit{\mathit{arrived}(\mathit{dest}})}$.
Furthermore, we can see that this is also a causal chain from~$\gamma(t_3)$ to this change in~$\mathcal{T}(\mathcal{S}[\gamma(t_3)])|_\gamma$.
Therefore, time\nobreakdash-step~$\gamma(t_3)$ is the unique candidate inflection point of change~${\mathit{\mathit{arrived}(\mathit{dest}})}$ and, thus, it is the unique inflection point as well.
As a result, singleton set~$\{{\bf do}(\mathit{approach},\gamma(t_3))\}$ is the unique cause of this change with respect to any such~$\gamma$.
Let us now consider an interpretation~$\gamma$ satisfying~$\gamma(t_4) \geq \gamma(t_3) + \gamma(\mathit{time2fork})$.
In this case $\mathit{arrived}(\mathit{dest})$ is still a change and~$P_1$ is the only proof of this change.
Hence,~$\{{\bf do}(\mathit{approach},\gamma(t_3))\}$ is also the unique cause of this change with respect to any such~$\gamma$.
Consequently, $\{{\bf do}(\mathit{approach},t_3)\}$ is the unique cause of this change in this story.

Let us now consider \emph{Suzy First} story (Example~\ref{ex:suzy.first}).
The unique answer set of~$\mathcal{T}_{\mathit{fst}}(\mathcal{S}_{\mathit{suzy}})|_\gamma$ contains atoms
\small
\begin{gather*}
  {\bf do}(a_1,(\gamma(t_1)),\, {\bf do}(a_2,\gamma(t_2)),\,
  \neg \mathit{broken}(0),\,
  \dotsc,
  \neg \mathit{broken}(n_4-1),\,
  \mathit{broken}(n_4)\,
\end{gather*}
\normalsize
with~$n_4 = \gamma(t_1)+\gamma(\mathit{duration}(a_1))$ being a positive integer representing the arriving time\nobreakdash-step of Suzy's rock.
This means that~$\mathit{broken}(n_4)$ is a change.
There is only one causal chain leading to this change:
\small
\begin{gather}
    \label{eq:causal.chain.suzy}
    {\bf do}(a_1,\gamma(t_1)),\, m_0(a_1),\,\mathit{broken}(n_4)
\end{gather}
\normalsize
and the only inflection point is~$\gamma(t_1)$.
As a result, Suzy's throw, $\{{\bf do}(a_1,t_1)\}$ is the only cause of this change.
Note that the order in which Suzy and Billy throw is irrelevant (as long as the constraint~$t_1 + \mathit{duration}(a_1) < t_2 + \mathit{duration}(a_2)$ is satisfied): the reason for Suzy's rock to get first may be because she throws first or because her rock is faster or any other reason.
It is easy to check that, if we consider a scenario where Billy's rock gets first -- formally a scenario~$\mathcal{S}_{\mathit{billy}}$ obtained from~$\mathcal{S}_{\mathit{suzy}}$ by replacing constraint~$t_1 + \mathit{duration}(a_1) < t_2 + \mathit{duration}(a_2)$ by~$t_1 + \mathit{duration}(a_1) > t_2 + \mathit{duration}(a_2)$ -- then Billy's throw, $\{{\bf do}(a_2,t_2)\}$, is the only cause of this change.

\noindent
In the following variation of \emph{Suzy First} story
$broken$ has two inflection points.
\begin{example}\label{ex:suzy.same}
  Suzy and Billy throw rocks at a bottle, but this time both rocks arrive at the same time.
\end{example}
This story can be formalized by a scenario~$\mathcal{S}_{\mathit{same}}$ obtained from~$\mathcal{S}_{\mathit{suzy}}$ by replacing
$t_1 + \mathit{duration}(a_1) < t_2 + \mathit{duration}(a_2)$ by $t_1 + \mathit{duration}(a_1) = t_2 + \mathit{duration}(a_2)$

For any interpretation~$\gamma$ of this scenario, we have change~$\mathit{broken}(n_5)$ with~$n_5 = \gamma(t_1) + \mathit{duration}(a_1) =\gamma(t_2) + \mathit{duration}(a_2)$ and two causal chains leading to this change:
\small
\begin{gather*}
    {\bf do}(a_1,\gamma(t_1)),\, m_0(a_1),\,\mathit{broken}(n_5)
\end{gather*}
\normalsize
\small
\begin{gather*}
    {\bf do}(a_2,\gamma(t_2)),\, m_0(a_2),\,\mathit{broken}(n_5)
\end{gather*}
\normalsize
Both~$\gamma(t_1)$ and~$\gamma(t_2)$ are inflection points. They may be the same inflection point or different ones depending of the interpretation~$\gamma$.
In all cases, both~$\{ {\bf do}(a_1,\gamma(t_1)) \}$ and~$\{ {\bf do}(a_2,\gamma(t_2)) \}$ are causes of change~$\mathit{broken}(n_5)$.
%

Let us now consider the variation of this story introduced in Example~\ref{ex:suzy.order}, where Suzy and Billy throw by the order of a stronger girl.
As we discuss above, $\mathit{broken}$ becomes true at time\nobreakdash-step~$t_1+\mathit{duration}(a_1)$.
In other words~$\mathit{broken}(n_4)$ is a change in scenario~$\mathcal{S}_{\mathit{order}}$.
%
In this scenario there is only one causal chain leading to this change:
\small
\begin{gather*}
    {\bf do}(b_1,0),\, m_6(a_1,t_1,b_1),\, m_0(a_1),\,\mathit{broken}(n_4)
\end{gather*}
\normalsize
and, thus, $\{{\bf do}(b_1,0)\}$ is the only cause of~$\mathit{broken}(n_4)$.

\noindent
Note that our notion of cause is different from the notion of immediate or direct cause.
The immediate cause of breaking the bottle is the throw of the rock, but the deliberate cause is the order.
We also discussed the scenario where both Suzy and Billy refuse to follow the order and, thus, $\mathit{broken}$ never happens.
Therefore, there was no cause.
Next let us consider a story where Suzy refuses to throw but Billy follows the order.
This can be formalized by scenario~$\mathcal{S}_{order3}$ obtained from~$\mathcal{S}_{order}$ by adding extended atom~${\bf do}(\neg a_1,t_1)$.
In this case the change happens later. That is, $\mathit{broken}(n_6)$ is a change with~$n_6 = \gamma(t_2)+\gamma(\mathit{duration}(a_2))$.
The only  causal chain leading to this change is
\small
\begin{gather*}
    {\bf do}(b_2,0),\, m_6(a_2,t_1,b_2),\, m_0(a_2),\, \mathit{broken}(n_6)
\end{gather*}
\normalsize
and, thus, $\{{\bf do}(b_2,0)\}$ is the only cause of~$\mathit{broken}(n_6)$.
Next example illustrates our treatment of preconditions of a cause.
%

\begin{example}
As in Example~\ref{ex:suzy.first} Suzy picks up a rock and throws it at the bottle.
However, this time we assume that she is accurate only if she aims first.
Otherwise, her rock misses.
Suzy aims before throwing and hits the bottle.
Billy just looks at his colleague's performance.
\end{example}
The story is formalized by causal theory ~$\mathcal{T}_{\mathit{aim}}$:
\small
\begin{gather*}
\begin{array}{rlll}
m_0'(A) : & \mathit{broken}(I) &\leftarrow &
         \mathit{occurs}(A,I-D),\mathit{member}(A,throw),\\
      &&& \mathit{agent}(A) = Ag, \mathit{duration}(A)=D,\\
     &&& \mathit{aimed}(Ag,I-D),\neg \mathit{broken}(I-1),\neg \mathit{ab}(m_0'(A),I)
\end{array}
\end{gather*}%
\normalsize
\small
\begin{gather*}
\begin{array}{rlll}
m_7(A) : & \mathit{aimed}(Ag,I) &\leftarrow &
         \mathit{occurs}(A,I-D),\mathit{member}(A,aim),\\
         &&& \mathit{agent}(A) = Ag, \mathit{duration}(A)=D,\neg \mathit{ab}(m_7(A),I)
\end{array}
\end{gather*}%
\normalsize
where $\mathit{aimed}$ is an inertial fluent and
scenario~$\mathcal{S}_{\mathit{aim}}$
%
\begin{flalign}
  \label{eq:aim.statics}
\begin{array}{l}
\mathit{agent}(a_1)=\mathit{suzy} \ \ \mathit{member}(a_1,throw)\ \
\mathit{duration}(a_1)\geq 1\\
\mathit{agent}(c) =\mathit{suzy} \ \ \ \ \mathit{member}(c,\mathit{aim}) \ \ \ \
  \ \ \mathit{duration}(c) \geq 1 \\
{\bf do}(c,t_5),\ {\bf
  do}(a_1,t_1), \ t_5 \!+\! \mathit{duration}(c) < t_1
\end{array}
\end{flalign}

The inflection point in $\mathcal{S}_{\mathit{aim}}$ is~$t_1$ and the only deliberate cause of $broken(n_4)$ is~$\{ {\bf do}(a_1,t_1) \}$.
Action~${\bf do}(c,t_5)$ is necessary for shattering the bottle,
because it is required by one of the preconditions of~$m_0^\prime(a_1)$.
However, it is not a deliberate cause because at the time of its occurrence the
shattering could not be predicted (see condition~(a) in Definition~\ref{candidateInflectionPoint}).

\st
Definition \ref{causeOfChange} can be used to define the notion of \emph{causal
  explanation} of unexpected observations:
 $T(S)$ is called \emph{strongly consistent} if $T^{reg}(S)$, obtained from
$T(S)$ by dropping cr-rules, is consistent.
If $T(S)$ is strongly consistent and $T(S \cup \{{\bf obs}(f,y,i)\})$
is not we say that ${\bf obs}(f,y,i)$ is
\emph{unexpected}. We assume that every abductive
support\footnote{\emph{Abductive support} of a program $\Pi$ is a minimal
  collection of cr-rules of $\Pi$ which, if turned into
  regular rules and added to the regular part of $\Pi$, produce a consistent program
  $\Pi^\prime$. Answer set of $\Pi$ is then defined as an answer set of
  $\Pi^\prime$. } $U$ of this theory has exactly one answer set.
By a cause of atom $f(i)=y$ we mean a cause of the last change of $f$ to
$y$ which precedes $i+1$ (note that for actions and time-independent fluent $f$, $f(i) = y$ is a change).
By \emph{causal explanation} of ${\bf
  obs}(f,y,i)$ we mean a cause of $f(i)=y$ in $T(S_U)$ where $S_U$ is obtained
from $S$ by adding ${\bf do}(a,i)$ for every rule $a(i) \rif $ from $U$ for some abductive support $U$. For example, consider a scenario $S$ of $\mathcal{T}_{\mathit{fst}}$
  consisting of ${\bf init}(\neg broken),{\bf obs}(broken, true,2)$,
  actions $a_1$ and $a_2$
  from $\mathcal{S}_{\mathit{suzy}}$ with durations $2$ and $4$ respectively.
The program has one abductive support, $a_1(0) \rif$ and hence ${\bf
  do}(a_1,0)$ explains  the unexpected
observation. If $broken$ were observed at $3$
we'd have two explanations: ${\bf do}(a_1,0)$ and ${\bf do}(a_1,1)$.
This can be compactly represented using a do-atom ${\bf do}(a_1,t)$ where $t$ is an abstract time step satisfying $0 \leq t < 2$.

\section{Conclusions}
The paper describes a new approach for representing causal knowledge, and its use for causal analysis.
The approach emphasizes the separation between background theory and scenario.
The first contains general knowledge that may be shared by different stories and the latter contains the information specific to the considered story.
This, together with the use of abstract constants, provides a higher degree of \emph{elaboration tolerance} than other approaches to causal analysis.
We also propose the use of a rich KR\nobreakdash-language that is able to represent sophisticated causal laws, time, defaults and their exceptions, recursive definitions, and other non-trivial phenomena of natural language.
As a result, we can obtain accurate and direct formalizations of natural
language sentences that, we believe, is essential for causal analysis.
We have illustrated this with common challenging examples from the literature on actual causality.
Causal analysis is realized over a formal representation rather than over the natural language statements.
However, our intuitions are usually more clear with respect to the natural language statements than with respect to the formal representation.
The closer the formal representation is to the natural language statements of a story, the better we can use our intuition to guide us towards a formal analysis of actual causality.
A preliminary version of this paper was presented at a workshop \cite{gelbal20a}.
%
We substantially extend that version and correct mistakes that were discovered after its presentation.
This has led us to change the definition of inflection point, to introduce the notion of tight proof and abstract time-steps,~etc.
In the future, this work should be expanded to consider other types of causal
relations. Some, like prevention, are not included due to
space limitation. Others require further work. In particular we plan to expand
$\mathcal{W}$ by probabilistic constructs of
P\nobreakdash-log 
and use it to study probabilistic causal relations.
%
%
%
%
 Finally, we plan to investigate mathematical properties of causal theories and
 algorithms for effectively computing the causes
of various causal relations and their implementations.
The notion of tight proof is closely related to the notion of
 causal justifications for answer set programs~\cite{cabfan14a}.
This may open the door to use \texttt{xclingo}~\cite{cafabr20a} as the first\nobreakdash-step of a new system for computing causes according to our definition.

\bibliographystyle{acmtrans}
\bibliography{causality}

\end{document}